\documentclass[10pt,twocolumn,letterpaper]{article}

\usepackage{cvpr}              

\definecolor{cvprblue}{rgb}{0.21,0.49,0.74}
\usepackage[pagebackref,breaklinks,colorlinks,allcolors=cvprblue]{hyperref}

\def\paperID{*****} 
\def\confName{CVPR}
\def\confYear{2025}

\title{PCIE\_Interaction Solution for Ego4D Social Interaction Challenge}

\author{
Kanokphan Lertniphonphan \\
Lenovo Research\\
{\tt\small klertniphonp@lenovo.com}
\and
Feng Chen\\
Lenovo Research\\
{\tt\small chenfeng13@lenovo.com}
\and
Junda Xu \\
Beijing Normal University\\
{\tt\small 202321081064@mail.bnu.edu.cn}
\and
Fengbu Lan \\
Tsinghua University\\
{\tt\small lanfb22@mails.tsinghua.edu.cn}
\and
Jun Xie \\
Lenovo Research\\
{\tt\small xiejun@lenovo.com}
\and
Tao Zhang \\
Tsinghua University\\
{\tt\small taozhang@tsinghua.edu.cn}
\and
Zhepeng Wang \\
Lenovo Research\\
{\tt\small wangzpb@lenovo.com}
}

\begin{document}
\maketitle
\begin{abstract}
This report presents our team's PCIE\_Interaction solution for the Ego4D Social Interaction Challenge at CVPR 2025, addressing both Looking At Me (LAM) and Talking To Me (TTM) tasks. The challenge requires accurate detection of social interactions between subjects and the camera wearer, with LAM relying exclusively on face crop sequences and TTM combining speaker face crops with synchronized audio segments. In the LAM track, we employ face quality enhancement and ensemble methods. For the TTM task, we extend visual interaction analysis by fusing audio and visual cues, weighted by a visual quality score. Our approach achieved 0.81 and 0.71 mean average precision (mAP) on the LAM and TTM challenges leader board. Code is available at \url{https://github.com/KanokphanL/PCIE_Ego4D_Social_Interaction}

\end{abstract}
    
\section{Introduction}
\label{sec:intro}

Egocentric video captured by wearable cameras has gained significant importance in computer vision and robotics. The Ego4D social interaction benchmark highlights interactions between the camera wearer and surrounding individuals through two key tasks: Looking At Me (LAM) and Talking To Me (TTM) \cite{Grauman2021Ego4DAT}.

The LAM task specifically detects whether a social partner's eye gaze is directed toward the camera wearer, utilizing face bounding boxes with cross-frame identity consistency. In contrast, the TTM task identifies conversational engagement through synchronized audio-visual analysis of tracked individuals. While both tasks employ frame-level predictions, TTM's labeling follows an utterance-level paradigm - even if the speaker momentarily looks away during a conversation segment, the interaction is still considered valid if the audio content indicates engagement with the camera wearer.

Several approaches \cite{Grauman2021Ego4DAT, gazepose} leverage image sequences to detect eye gaze toward the camera wearer by analyzing visual cues from face crops. \cite{gazepose} utilizes facial landmarks and head pose estimation to enhance spatial feature refinement. \cite{Grauman2021Ego4DAT} incorporates both spatial and temporal dynamic refinement. More recently, \cite{pcie2024lam} introduced a framework combining the InternVL image encoder \cite{chen2023internvl} with a Bi-LSTM network to jointly model spatial and temporal dependencies.

The baseline model for TTM \cite{Grauman2021Ego4DAT} extends the LAM architecture by incorporating an audio encoder to extract embeddings from synchronized audio segments. These audio embeddings are concatenated with visual features and jointly classified to predict TTM interactions. However, \cite{Lin2023QuAVFQA} demonstrated that training separate models for audio and vision can gain a better results. Their method processes audio and visual modalities independently, then fuses the predictions using a face quality score to weight the TTM score of each modality.

Our approach for social interaction analysis extends the LAM to the TTM task by employing separate models for visual and audio modalities. Based on the dataset characteristic, we introduce visual and audio filter before fusing the score from both models, yielding a robust final interaction score for TTM.

\section{Method}
\label{sec:method}

\begin{figure*}[t]
  \centering
   \includegraphics[width=1\linewidth]{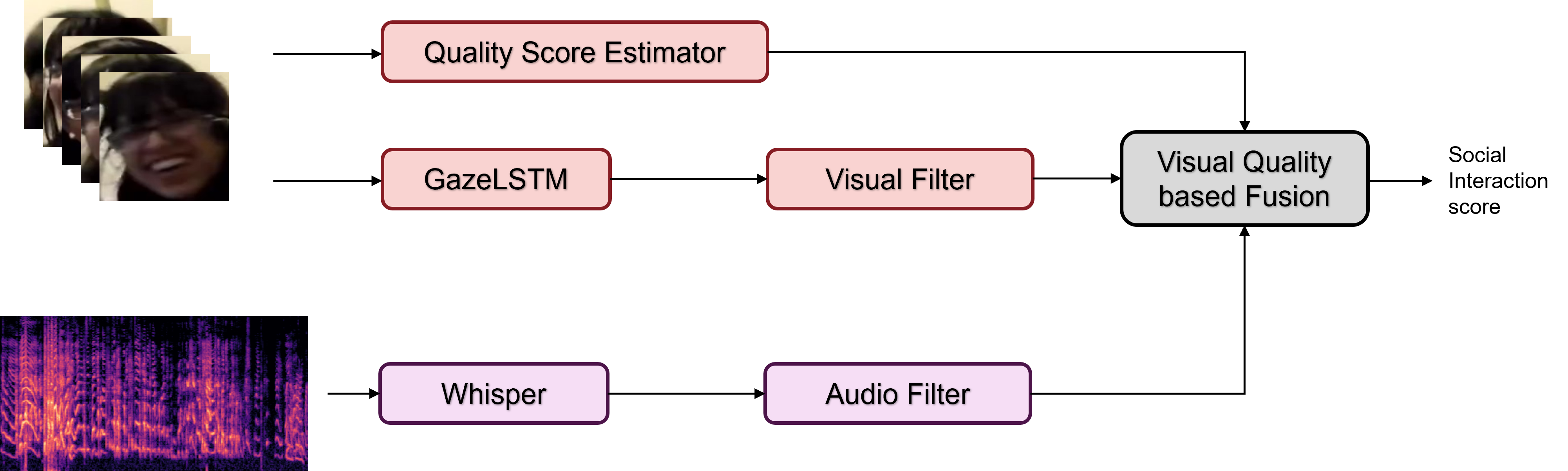}
   \caption{
   Our proposed Social Interaction Detection framework  
   }
   \label{fig:framework}
\end{figure*}

Our proposed social interaction framework for the Ego4D challenges, illustrated in Figure \ref{fig:framework}, addresses both Looking At Me (LAM) and Talking To Me (TTM) tasks through a designed multimodal architecture. The system processes visual and audio modalities through independent, task-optimized branches, with each stream passing through modality-specific filters prior to fusion. For LAM, we employ only the visual branch to detect gaze patterns. The TTM solution incorporates both modalities: visual predictions undergo max-score filtering to generate utterance-level representations, which are then fused with processed audio features through a quality-weighted fusion module. This dual-stream approach allows each modality to compensate for the other's limitations while maintaining task-specific optimization.

\subsection{Looking At Me}

The LAM dataset's face crops exhibit significant motion blur due to relative movement between wearable cameras and subjects. Our analysis of GazeLSTM \cite{Grauman2021Ego4DAT} demonstrated that augmenting the architecture with a transformer decoder yielded minimal improvement on validation data. Drawing from established methods, we developed an ensemble framework that integrates multiple model outputs while addressing image quality by applying deblurring techniques \cite{chu2022nafssr} to enhance input face crops. 

\begin{table*}
    \centering
    \begin{tabular*}{0.8\textwidth}{@{\extracolsep{\fill}}lcccc} 
    \toprule
         &  \multicolumn{2}{c}{Validation}  &  \multicolumn{2}{c}{Test} \\
        \hline
        \multicolumn{1}{c}{Method}  & mAP ($\uparrow$) & Acc ($\uparrow$) & mAP ($\uparrow$) & Acc ($\uparrow$)  \\
        \midrule
        GazeLSTM \cite{Grauman2021Ego4DAT} & 79.40\% & 92.78\% & 74.54\% & 87.04\% \\
        GazeLSTM with Transformer head & 76.81\% & 91.66\% & 73.29\% & 87.09\% \\
        GazePose \cite{gazepose} & 81.89\% & 91.84\% & 76.95\% & 85.08\% \\
        InternLSTM \cite{pcie2024lam} & 84.23\% & 94.79\% & 76.57\% & 90.00\%\\
        \bottomrule
  \end{tabular*}
  \caption{Looking at me experimental results}
  \label{tab:base}
\end{table*}

\subsection{Talking To Me}

We adopt a dual-modality approach for the TTM task, implementing separate visual and audio models following \cite{Lin2023QuAVFQA}. For audio processing, we employ the Whisper encoder \cite{radford2022whisper}, evaluating both small and large-v3 architectures. While the large model performs better on validation data (Table \ref{tab:talk1}), this improvement does not generalize to the test set, suggesting potential overfitting to the validation distribution.

The TTM annotation scheme presents unique challenges for visual modality integration. Since labels are assigned at the utterance level based on audio segments, speakers may intermittently look away while still engaged in conversation or might not look at the camera wearer at all. To address this mismatch between frame-level visual predictions and utterance-level labels, we implement a visual max-score filter. This filter aggregates visual predictions across each audio segment, retaining only the maximum visual score as the utterance-level representation. As shown in Table \ref{tab:talk1}, this filtering significantly improves visual-only performance.

For multimodal fusion, we compare two approaches: (1) basic score averaging and (2) quality-weighted fusion \cite{Lin2023QuAVFQA}, where weights are derived from frame-level face alignment scores \cite{bulat2017far}. Final predictions undergo median filtering to smooth temporal inconsistencies before submission.

\begin{table*}
    \centering
    \begin{tabular*}{0.8\textwidth}{@{\extracolsep{\fill}}lcccc} 
    \toprule
         &  \multicolumn{2}{c}{Validation}  &  \multicolumn{2}{c}{Test} \\
        \hline
        \multicolumn{1}{c}{Method}  & mAP ($\uparrow$) & Acc ($\uparrow$) & mAP ($\uparrow$) & Acc ($\uparrow$)  \\
        \midrule
        \textbf{Audio modality}\\
        Whisper\_small  & 70.17\% & 63.91\% & 66.16\% & 60.80\% \\
        Whisper\_large\_v3 & 70.51\% & 72.52\% & 66.17\% & 56.92\% \\
        \hline
        \textbf{Visual modality}\\
        GazeLSTM & 63.75\% & 71.24\% & 54.86\% & 53.86\% \\
        + Visual filter &  &  & 63.27\% & 59.57\%\\
        \hline
        \textbf{Audio and  Visual modality}\\
        Averaging &  &  & 68.78\% & 62.32\%\\
        Quality Score weight Fusion &  &  & 69.81\% & 58.56\%\\
        \bottomrule
  \end{tabular*}
  \caption{Talking to me experimental results}
  \label{tab:talk1}
\end{table*}

\section{Experiments}
\label{sec:experiment}

\subsection{Dataset and Evaluation metric}

We evaluate our approach on the Ego4D social benchmark \cite{Grauman2021Ego4DAT}, which contains 572 video clips (5 minutes each), split into 389 training, 50 validation, and 133 test clips. The benchmark provides two distinct annotation schemes. For LAM, frame-level binary labels indicating whether each detected face is looking at the camera wearer. For TTM, utterance-level labels for audio-visual segments, marking whether the speaker is talking to the camera wearer. Mean Average Precision (mAP) and Top-1 accuracy are used for evaluation in both tasks.

\subsection{Results}

\textbf{Looking At Me}

We first reproduced the baseline approach from \cite{Grauman2021Ego4DAT}, extending its architecture with a transformer decoder. For comprehensive comparison, we also implemented two state-of-the-art methods: GazePose \cite{gazepose} and InternLSTM \cite{pcie2024lam} (see Table \ref{tab:base}). To enhance input quality, all test face crops were preprocessed using the image enhancement module to reduce motion blur and improve image quality.

\begin{figure*}[t]
  \centering
   \includegraphics[width=1\linewidth]{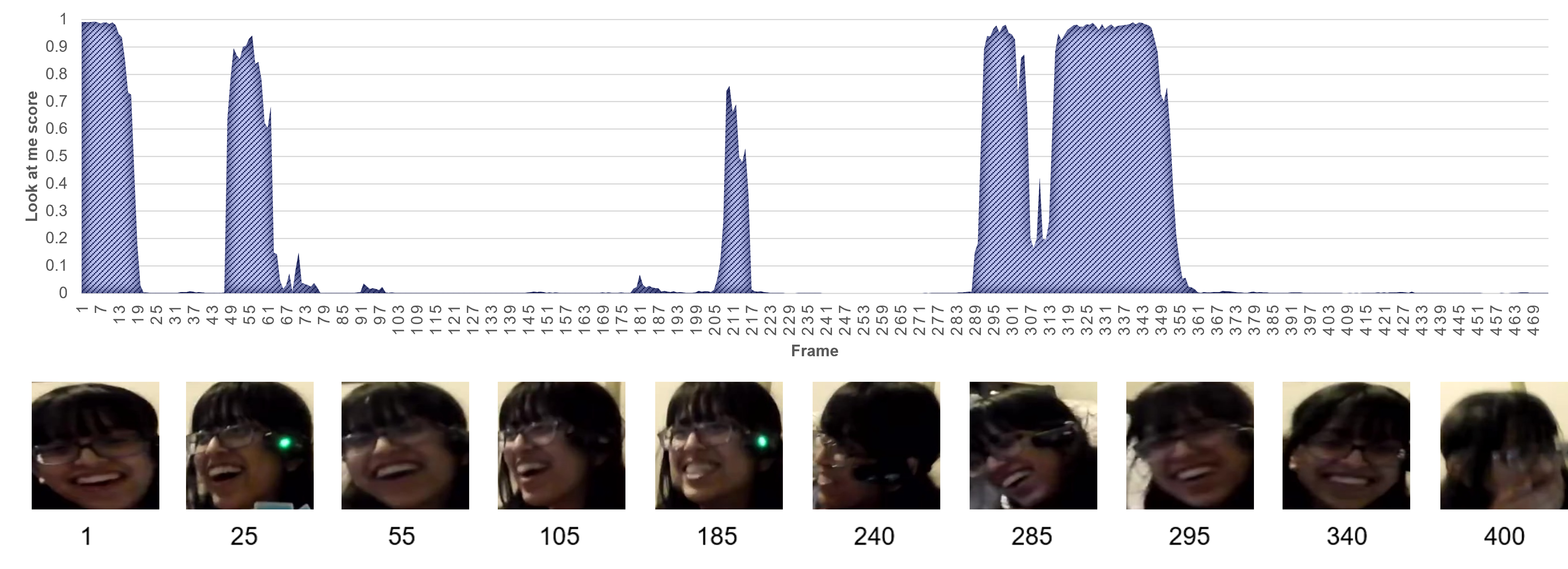}
   \caption{
   Looking At Me results   
   }
   \label{fig:lam_pic}
\end{figure*}

Our final submission combines predictions from multiple approaches through an ensemble of InternLSTM, GazeLSTM, and GazePose models with diverse parameter configurations. We applied a median filter to the ensemble outputs for temporal smoothing, achieving robust performance. This approach attained top results on the Ego4D Looking At Me benchmark with a mean Average Precision (mAP) of 0.81 and Top-1 accuracy of 0.93, as shown in Table \ref{tab:Lam_final}. The results for LAM through the image sequence are shown in Fig \ref{fig:lam_pic}

\begin{table}
  \centering
  \begin{tabular}{l c c}
    \toprule
    \multicolumn{1}{c}{Team} & mAP ($\uparrow$) & Acc ($\uparrow$)  \\
    \midrule
    PCIE\_LAM (Ours) & \textbf{0.81} & \textbf{0.93} \\
    USTB\_LAM & 0.80 & 0.87 \\
    PKU-WICT-MIPL & 0.79 & 0.92 \\
    ydejie & 0.79 & 0.92 \\
    Dejie (gb6) & 0.79 & 0.92 \\
    Host\_24191\_Team & 0.66 & 0.74 \\ 
    \bottomrule
  \end{tabular}
  \caption{Ego4D Looking at me Challenge Leaderboard}
  \label{tab:Lam_final}
\end{table}

\bigbreak
\textbf{Talking To Me}

The Whisper-based audio feature extraction demonstrates strong sensitivity to conversational speech, producing consistently high confidence scores for clear dialogue segments. However, we found that the model shows lower output scores for non-standard vocal expressions (e.g., exclamations or laughter). This pattern results in an elevated false positive rate by using only audio modality.

The visual modality presents unique challenges compared to LAM, as subjects frequently avert their gaze during conversations (Figure \ref{fig:lam_pic}). In addition, the label of conversation with a group of people is 1 or talking to me. To handle with TTM task, we pass the prediction score to the visual max-score filter to align the visual results with the TTM protocol. By applying the filter, the testing result of the visual modality improves significantly and is comparable with audio modality results as shown in Table \ref{tab:talk1}. 

While audio effectively detects verbal engagement, visual cues provide crucial disambiguation for non-interactive scenarios. This synergy proves particularly valuable for filtering false positives from the audio-only modality, where conversation detection does not necessarily indicate social interaction. For our final submission, we ensemble multiple fusion strategies and apply temporal post-processing to optimize consistency. Our submission result on the TAM leader board is shown in Table \ref{tab:ttm_final}  

\begin{table}
  \centering
  \begin{tabular}{l c c}
    \toprule
    \multicolumn{1}{c}{Team} & mAP ($\uparrow$) & Acc ($\uparrow$)  \\
    \midrule
    PCIE\_LAM (Ours) & \textbf{0.71} & \textbf{0.60} \\
    ex2eg & 0.68 & 0.61 \\
    z(test) & 0.67 & 0.62 \\
    HsiCheLin & 0.67 & 0.58 \\
    EgoAdapt & 0.67 & 0.62 \\
    Host\_30426\_Team & 0.54 & 0.54 \\ 
    \bottomrule
  \end{tabular}
  \caption{Ego4D Talking to me Challenge Leaderboard}
  \label{tab:ttm_final}
\end{table}


\section{Conclusion}

We propose a unified framework for social interaction detection in the Ego4D challenge, addressing both Looking At Me (LAM) and Talking To Me (TTM) tasks through optimized multimodal processing. Our approach employs independently trained visual and audio models to maximize modality-specific performance. For LAM, we ensemble models with diverse parameter configurations, while TTM processing incorporates visual max-score filtering to align frame-level gaze predictions with utterance-level audio segments. The framework demonstrates effective complementary fusion: while audio reliably detects verbal engagement, visual cues crucially eliminate false positives from non-interactive scenarios, indicating better performance than unimodal approaches. 
{
    \small
    \bibliographystyle{ieeenat_fullname}
    \bibliography{main}

\begin{thebibliography}{8}
\providecommand{\natexlab}[1]{#1}
\providecommand{\url}[1]{\texttt{#1}}
\expandafter\ifx\csname urlstyle\endcsname\relax
  \providecommand{\doi}[1]{doi: #1}\else
  \providecommand{\doi}{doi: \begingroup \urlstyle{rm}\Url}\fi

\bibitem[Bulat and Tzimiropoulos(2017)]{bulat2017far}
Adrian Bulat and Georgios Tzimiropoulos.
\newblock How far are we from solving the 2d \& 3d face alignment problem? (and a dataset of 230,000 3d facial landmarks).
\newblock In \emph{International Conference on Computer Vision}, 2017.

\bibitem[Chen et~al.(2023)Chen, Wu, Wang, Su, Chen, Xing, Zhong, Zhang, Zhu, Lu, Li, Luo, Lu, Qiao, and Dai]{chen2023internvl}
Zhe Chen, Jiannan Wu, Wenhai Wang, Weijie Su, Guo Chen, Sen Xing, Muyan Zhong, Qinglong Zhang, Xizhou Zhu, Lewei Lu, Bin Li, Ping Luo, Tong Lu, Yu Qiao, and Jifeng Dai.
\newblock Internvl: Scaling up vision foundation models and aligning for generic visual-linguistic tasks.
\newblock \emph{arXiv preprint arXiv:2312.14238}, 2023.

\bibitem[Chu et~al.(2022)Chu, Chen, and Yu]{chu2022nafssr}
Xiaojie Chu, Liangyu Chen, and Wenqing Yu.
\newblock Nafssr: Stereo image super-resolution using nafnet.
\newblock In \emph{Proceedings of the IEEE/CVF Conference on Computer Vision and Pattern Recognition (CVPR) Workshops}, pages 1239--1248, 2022.

\bibitem[et~al.(2021)]{Grauman2021Ego4DAT}
Kristen~Grauman et al.
\newblock Ego4d: Around the world in 3,000 hours of egocentric video.
\newblock \emph{2022 IEEE/CVF Conference on Computer Vision and Pattern Recognition (CVPR)}, pages 18973--18990, 2021.

\bibitem[Lertniphonphan et~al.(2024)Lertniphonphan, Xie, Meng, Wang, Chen, and Wang]{pcie2024lam}
Kanokphan Lertniphonphan, Jun Xie, Yaqing Meng, Shijing Wang, Feng Chen, and Zhepeng Wang.
\newblock Pcie\_lam solution for ego4d looking at me challenge.
\newblock \emph{ArXiv}, abs/2406.12211, 2024.

\bibitem[Lin et~al.(2023)Lin, Wang, Chen, Fu, and Wang]{Lin2023QuAVFQA}
Hsi-Che Lin, Chien-Yi Wang, Min-Hung Chen, Szu-Wei Fu, and Y. Wang.
\newblock Quavf: Quality-aware audio-visual fusion for ego4d talking to me challenge.
\newblock \emph{ArXiv}, abs/2306.17404, 2023.

\bibitem[Radford et~al.(2022)Radford, Kim, Xu, Brockman, McLeavey, and Sutskever]{radford2022whisper}
Alec Radford, Jong~Wook Kim, Tao Xu, Greg Brockman, Christine McLeavey, and Ilya Sutskever.
\newblock Robust speech recognition via large-scale weak supervision, 2022.

\bibitem[Wei et~al.()Wei, Yang, Peng, and Liu]{gazepose}
Xiyu Wei, Dejie Yang, Yuxin Peng, and Yang Liu.
\newblock Team pku-wict-mipl ego4d look at me challenge 2023 technical report.

\end{thebibliography}
}


\end{document}